\documentclass[a4paper, 10pt, conference]{svproc}

\usepackage{amsmath} 
\usepackage{amssymb}  
\usepackage{graphicx}
\usepackage{multicol}
\usepackage{footmisc}
\usepackage[table,xcdraw]{xcolor}
\usepackage{booktabs}
\usepackage{caption}
\usepackage{subcaption}
\usepackage{tabulary}
\usepackage{multirow}
\PassOptionsToPackage{hyphens}{url}\usepackage{hyperref}

\def\figref#1{Figure~\ref{#1}}

\def\secref#1{Section~\ref{#1}}

\def\eqref#1{equation~\ref{#1}}

\newcommand{\xxnote}[3]{}
\ifx\hidenotes\undefined
  \usepackage{color}
  \renewcommand{\xxnote}[3]{\color{#2}{#1: #3}}
\fi

\begin{document}
\mainmatter

\title{
Robot-Assisted Feeding: Generalizing Skewering Strategies across Food Items on a Plate
}
\titlerunning{Generalizing Skewering Strategies}
\author{Ryan Feng*, Youngsun Kim*, Gilwoo Lee*, Ethan K. Gordon, Matt Schmittle, Shivaum Kumar{$^\dagger$}, Tapomayukh Bhattacharjee, Siddhartha S. Srinivasa
\thanks{*These authors contributed equally to the work.}}
\authorrunning{Feng et al.}
\institute{Paul G. Allen School of Computer Science \& Engineering, University of Washington, 185 E Stevens Way NE, Seattle, WA, USA\\\email \{rfeng, yskim, gilwoo, ekgordon, schmttle, tapo, siddh\}@cs.uw.edu, {$^\dagger$}kumars7@uw.edu}

\maketitle

\begin{abstract}
A robot-assisted feeding system must successfully acquire many different food items. 
A key challenge is the wide variation in the physical properties of food, demanding diverse acquisition strategies that are also capable of adapting to previously unseen items. Our key insight is that items with similar physical properties will exhibit similar success rates across an action space, allowing the robot to generalize its actions to previously unseen items. To better understand which skewering strategy works best for each food item, we collected a dataset of 2450 robot bite acquisition trials for 16 food items with varying properties. Analyzing the dataset provided insights into how the food items' surrounding environment, fork pitch, and fork roll angles affect bite acquisition success. We then developed a bite acquisition framework that takes the image of a full plate as an input, segments it into food items, and then applies our Skewering-Position-Action network (SPANet) to choose a target food item and a corresponding action so that the bite acquisition success rate is maximized. 
SPANet also uses the surrounding environment features of food items to predict action success rates. We used this framework to perform multiple experiments on uncluttered and cluttered plates.
Results indicate that our integrated system can successfully generalize skewering strategies to many previously unseen food items.


\end{abstract}

\section{Introduction}
 Eating is a vital activity of daily living at home or in a community. Losing the ability to feed oneself can be devastating to one's sense of self-efficacy and autonomy~\cite{jacobsson2000people}. Helping the approximately 1.0 million US adults who require assistance to eat independently~\cite{brault2012americans} would improve their self-worth~\cite{prior1990electric,stanger1994devices}.  It would also considerably reduce caregiver hours since feeding is one of a caregiver's most time-consuming tasks~\cite{kayser-jones1997staffingQualityCare,chio2006caregiver}.

According to a taxonomy of manipulation strategies developed for the feeding task~\cite{bhattacharjee2018food}, feeding requires the \textit{acquisition} of food items from a plate or bowl and the \textit{transfer} of these items to a person. 
In this paper, we address the challenge of generalization to previously unseen food items during the \textit{bite acquisition phase} of the feeding task.
Bite acquisition requires the perception of food items on a cluttered plate and the manipulation of these deformable items, which is challenging for two reasons. First, the universe of food items is immense, so we cannot expect to to have previously trained on every type of food item. Second, bite acquisition involves complex and intricate manipulation to handle food with a variety of physical characteristics, such as sizes, shapes, compliance, textures.

Our previous work~\cite{gallenberger2019transfer} showed that one can train a classifier to identify where and how to skewer an item.
While promising, the proposed approach requires classifying food items and manually labeling skewering locations for each food type, which does not generalize to unseen food items.

Our key insight is that items with similar physical properties (e.g., shapes and sizes) will often exhibit similar success rates when the same action is used to acquire them. If so, a robot can apply an action that was successful for acquiring a known food item to unknown items of similar properties. 

\begin{figure*}[t!]
\centering
\begin{subfigure}[b]{0.49\columnwidth}
    \includegraphics[width=\linewidth]{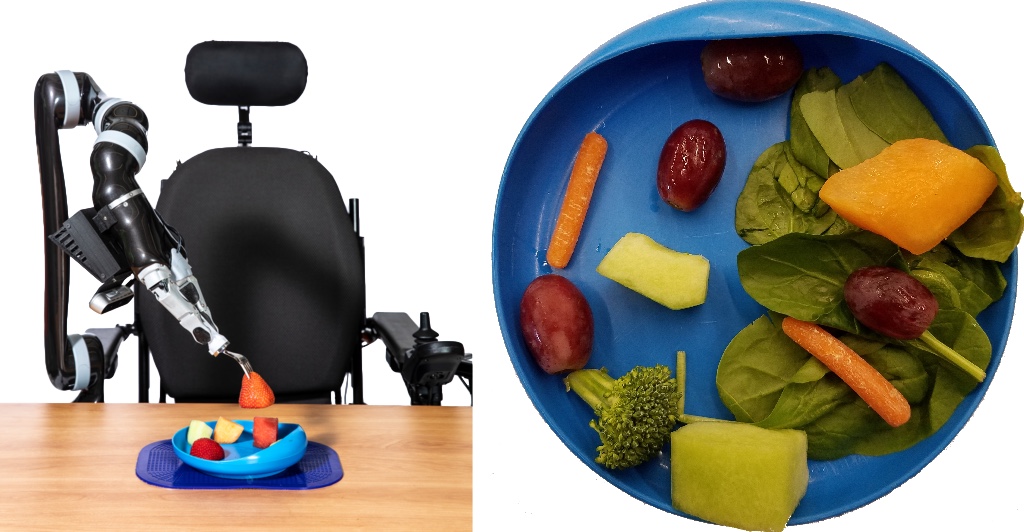}
    \label{fig:motivation}
\end{subfigure}
\hfill
\begin{subfigure}[b]{0.49\columnwidth}
    \includegraphics[width=\linewidth]{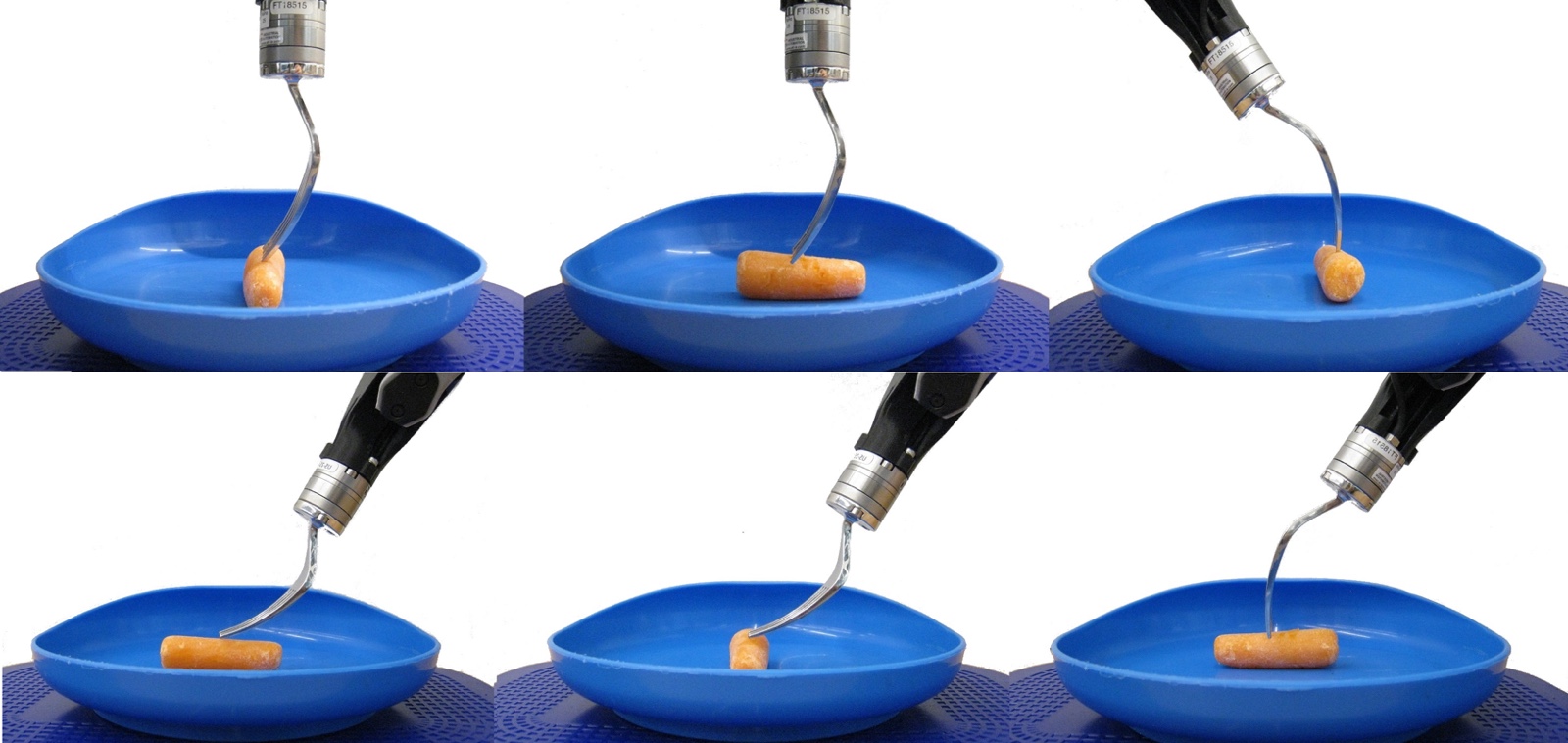}
    \label{fig:schematic_subfig}
    \end{subfigure}
 \caption{\emph{Left:} A robot acquiring a food item from a plate. \emph{Center:} A plate cluttered with food items. Food items can be isolated, near a wall (i.e., the edge of a plate) or another food item, or stacked on top of other food items. \emph{Right:} Three macro actions: VS: vertical skewer (\emph{two in the upper left}), TV: tilted skewer with vertical tines (\emph{two on the right}), TA: tilted skewer with angled tines (\emph{two in the lower left}). Each macro action has two fork rolls ($0 ^\circ$ and $90 ^\circ$).}
  \vspace{-0.5cm}
 \label{fig:schematic}
\end{figure*}

Our insight led us to a data-driven approach for autonomous bite acquisition that generalizes to previously unseen food items.
We collected a large dataset of $2450$ bite acquisition trials for $16$ food items using $6$ actions. 
Using this dataset, we trained a classifier to directly predict success rates of the six actions given a cropped image of a food item and its environmental features that we empirically found to be important when choosing an action. 
We refer to this model as the \emph{Skewering-Position-Action Network} (SPANet).

Our analysis shows that SPANet learns to predict success rates accurately for trained food classes and generalizes well to unseen food classes. Our final integrated system uses these predicted success rates to acquire various foods items from cluttered and non-cluttered plates (see \figref{fig:overview}).

Our contributions include:
\begin{itemize}
    \item \textbf{A dataset} of bite acquisition trials with success rates for food items with various physical properties and in varied environments~\cite{DVN/8A1XO3_2019}
    \item \textbf{An algorithm} that can generalize bite acquisition strategies to unseen items
    \item \textbf{A framework} for bite acquisition from a plate with various food items
\end{itemize}

\section{Related Work}

\subsection{Food Manipulation for Assistive Feeding}
Manipulating food for assistive feeding poses unique challenges compared to manipulating it for other applications~\cite{gemici2014learning,beetz2011pancakes}. Previous studies have created a food item taxonomy, explored the role of the haptic modality~\cite{bhattacharjee2018food}, developed scooping strategies~\cite{park2019toward,kobayashi2016meal}, and proposed an algorithm for intelligently selecting a strategy from a set of expert-defined skewering strategies~\cite{gallenberger2019transfer}. While expert-defined strategies achieve a good degree of success~\cite{gallenberger2019transfer}, they require item classification and therefore do not generalize to previously unseen food items. In this paper, our goal is to develop a method that generalizes acquisition strategies to previously unseen food items. 

\begin{figure*}[t!]
\includegraphics[width=\linewidth]{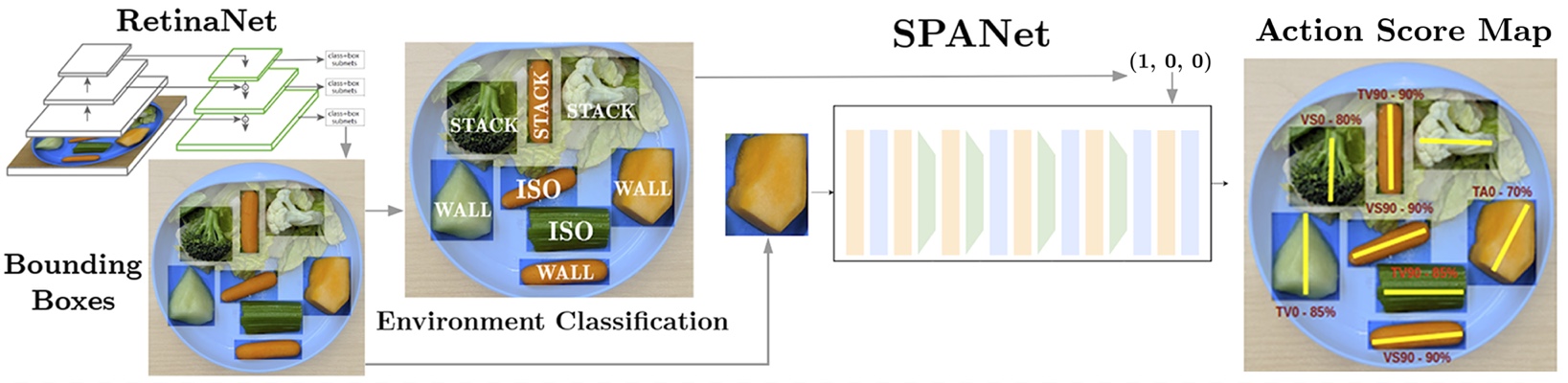}
 \caption{Our framework: Using a full-plate image, RetinaNet outputs bounding boxes around food items. An environment classifier identifies items as being in one of three environments: isolated (ISO), near a wall (i.e., plate edge) or another food item (WALL), or on top of other food items (STACK). SPANet uses the bounding boxes and  environment features to output the \textit{success probability for each bite acquisition action} and the \textit{skewering axis} for each food item.}
 \label{fig:overview}
\end{figure*}

Food manipulation for assistive feeding shares similarities to existing literature on grasping. Determining a good grasping position resembles finding a good skewering location. One approach in grasping literature uses learning-based methods to identify good grasping locations on a 3d model and draws them out to perceiving objects~\cite{mahler2017dex}. Others use real robot grasping trials~\cite{pinto2016graspingSelfSupervised} or images~\cite{saxena2008graspingNovelObjectsVision,redmon2015graspDetectionCNN}. However, these approaches generally focus on rigid objects, while our work requires tool-mediated, non-prehensile manipulation of deformable objects for which the haptic modality plays a crucial role. Importantly, physical properties of a food item may change after a failed skewering attempt, underlying the importance of an intelligent first action.

Our work is most closely related to our previous work~\cite{gallenberger2019transfer} which trains a classifier, SPNet ({Skewering-Position Network}), to identify a food item and propose skewering positions. The food classification is used to determine the macro skewering strategy, i.e., tilted or vertical skewering actions.
In total, SPNet proposes one of $5202$ discrete skewering actions, while our new method predicts success rates for only 6 actions that target the center of a food item in different directions~(\figref{fig:schematic}).
The main reason for this reduced action space is for generalization to unseen food items, for which highly targeted skewering position may not transfer.

In addition, unlike SPNet~\cite{gallenberger2019transfer} which relied on human labels of skewering positions, we use empirical success rates from robot-driven bite acquisition trials, which is a more accurate measure of the robot's performance.

\subsection{Food Perception}
Many strides in food perception have been made, especially with respect to deep supervised learning using computer vision systems, e.g., for food detection~\cite{hamid2016foodRecognition,ashutosh2016foodClassification,yanai2015foodRecognition}. Our previous work~\cite{gallenberger2019transfer} uses RetinaNet as an efficient, two-stage food detector that is faster than other two-stage detectors but more accurate than single-shot algorithms. The generated food item bounding boxes are then input to our SPNet~\cite{gallenberger2019transfer}, which generates position and rotation masks for skewering those food items. This idea was adopted from \cite{He2018-ec}, which uses independent masks from Mask R-CNN with Fully Convolutional Network (FCN) branches. 

An important question for the generalization of food items concerns out-of-class classification in perception problems. Since it is not feasible to train on every possible food item, systems like ours must intelligently handle items missing from the training set without being reduced to taking random actions. This work is related to the general task of detecting out-of-class examples. One common baseline approach uses a softmax confidence threshold~\cite{hendrycks17baseline} to detect such examples. Another looks at projecting seen and unseen categories into a joint label space and relating similar classes by generalizing off of zero-shot learning methods~\cite{chao2016empirical}.
Similarly, we aim to generalize our actions to unseen food items based on ones with similar properties.

\section{Bite Acquisition on a Plate}
Autonomous feeding system should take into account many factors such as bite acquisition, user preferences, and bite sequence. In this work, we focus on bite acquisition, and aim to maximize the success rate of a single item bite acquisition. We call a bite acquisition attempt a success if the robot picks up a target item.

Towards that end, we design an action selector that chooses an action and a target given the RGBD image of a full plate such that the success rate is maximized.
For each food item, we define the \emph{skewering axis} as the major (longer) axis of the item of a fitted ellipse. Approach angles and orientations for all actions were defined with respect to the center of the skewering axis, with $0$ degrees being parallel to the skewering axis.

Each action is implemented as a closed-loop controller that takes as input multimodal information (i.e., RGBD images and haptic signals) and outputs a low-level control (e.g., fork-tine movement). The haptic signal, given by a force/torque sensor attached to our fork, is not used as an input feature to our model (SPANet) but is used to detect if a force threshold is met during skewering.

\subsection{Action Space}\label{ssec:actions}
Three macro actions are based on the angle of approach (fork pitch). For each macro action, we have two fork rolls at $0$ and $90$ degrees, where $0$ degree means the fork tines are parallel to the skewering axis (see \figref{fig:overview} right), which leads to a total of six actions. These actions are derived from our previous work~\cite{gallenberger2019transfer}.

\noindent \textbf{Vertical Skewer (VS)}: The robot moves the fork over the food so the \textit{fork handle} is vertical, moves straight down into the food, and moves straight up. 

\noindent \textbf{Tilted Vertical Skewer (TV)}: The robot tilts the fork handle so that the \textit{fork tines} are vertical, creating a vertical force straight into the food item. The inspiration for this action comes from foods such as grapes, where a straight vertical force could prevent the grapes from rolling away.

\noindent \textbf{Tilted Angled Skewer (TA)}: The robot 
diagonally approaches the food with horizontal fork tines. The inspiration for this action comes from foods such as bananas, which are soft and require support from the fork to be lifted~\cite{bhattacharjee2018food}.

\begin{figure*}[t!]
\centering
\includegraphics[width=\linewidth]{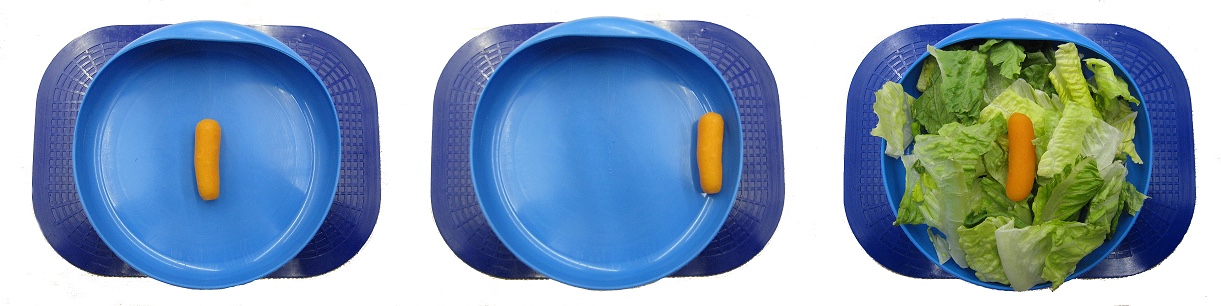}
 \caption{Three \emph{environment features}: Isolated (\emph{left}), Wall (\emph{middle}), and Stack (\emph{right}) scenarios. Note, the wall scenario can also be triggered if the desired food item is near another food item.}
 \label{fig:schematic2}
\end{figure*}

\subsection{Environment Features}
Three environment features are used in our model. They showcase the properties of the immediate surrounding environment of a food item (see \figref{fig:schematic2}).

\noindent \textbf{Isolated}: The surrounding environment of the target food item is empty.
This scenario may arise on any cluttered plate and becomes more common as the feeding task progresses and the plate empties.

\noindent \textbf{Wall}: The target food item is either close to the edge (wall) of a plate or bowl or close to another food item that acts as a support. This scenario may arise in any cluttered plate configuration.

\noindent \textbf{Stack}: The target food item is stacked on top of another food item. This scenario may arise in any cluttered plate configuration (e.g., a salad). To mimic food plate setups similar to a salad plate, we used lettuce as an item on which to stack other food items. In this work, we do not consider food items stacked on items other than lettuce.

\section{Generalizing to Previously Unseen Food Items}
We now develop a model for predicting bite acquisition success rates that can potentially generalize to previously unseen food items. In our prior study~\cite{gallenberger2019transfer}, we built a model that finds the best vertical skewering position and angle from the input RGB image; this model showed high performance for trained food items. However, identity information served a significant role in the detection model, so the model did not generalize to unseen food items even if they could be skewered with the same action. We also used a hand-labeled dataset for where to skewer, which reflected subjective annotator intuitions.

To build a generalizable model that is not biased by human intuition, we propose three solutions. First, we increase the range of food items compared to ~\cite{gallenberger2019transfer}. Second, we use empirical success rates based on real-world bite acquisition trials performed by a robot instead of manual labels. Third, we directly predict success rates over our action space instead of predicting the object class to select its predefined action. Our previous model~\cite{gallenberger2019transfer} output a $17 \times 17$ grid of fork positions and one of 18 discretized fork roll angles, leading to an action space of 5202 discrete actions. In this work, we developed a model that abstracts the position and rotation by using a skewering axis, reducing the action space to just 6 actions, which could improve the model's generalizability. With this approach, regression for the skewering axis is more general and easier to solve, and the quantity and variety of data required for training can be significantly reduced.

\begin{figure*}[t!]
\centering
\includegraphics[width=\linewidth]{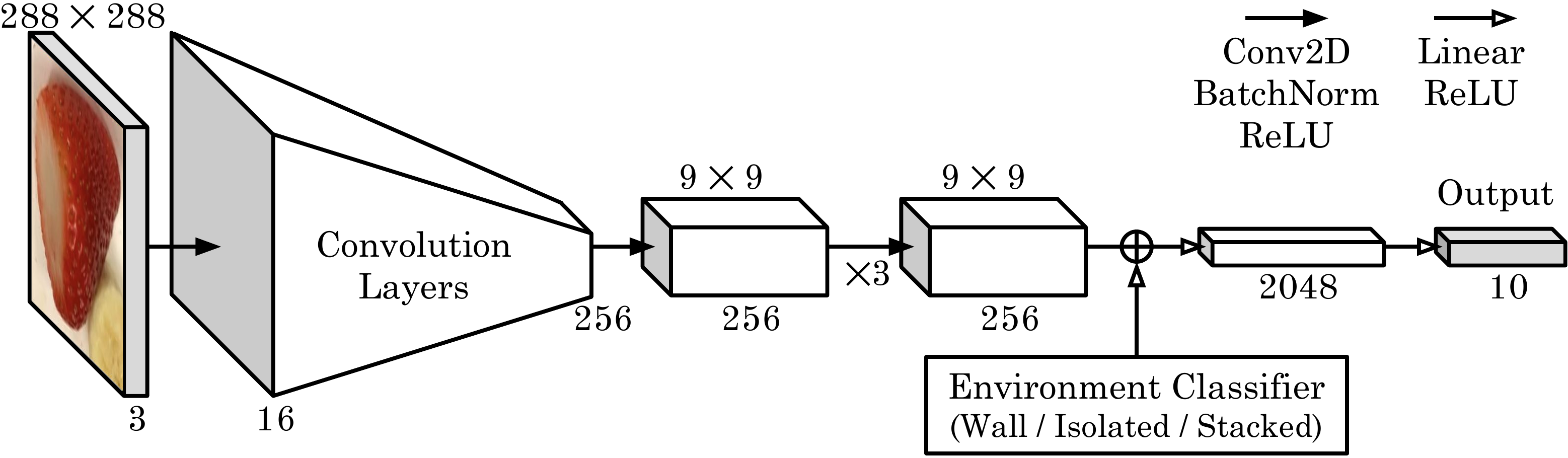}
 \caption{The architecture of SPANet, which predicts success rates of the six actions and the two end points of the major axis.}
 \label{fig:network_architecture}
\end{figure*}

\subsection{Skewering-Position-Action Network: SPANet}
We designed a network to visually determine the location to skewer and the acquisition strategy to take.
Our network receives a $288 \times 288 \times 3$ RGB image from a wrist-mounted camera as input; it outputs a $10 \times 1$ vector containing the two endpoints of the main axis and $k$ predicted success rates, where $k = 6$ to depict the three macro action types over two fork roll angles. 

We experimented with two network architecture designs. The first used a DenseNet~\cite{Huang2016-wv} base that was pre-trained on ImageNet~\cite{Russakovsky2014-tg}. The second used an AlexNet~\cite{Krizhevsky2012-qz} style base with simple convolution layers with batch normalization. In both cases, the base network was followed by three fully convolutional layers and a linear layer for the final output.
We decided upon using the latter, which ran 50\% faster with only a marginal performance hit. Even with $7$ convolutional layers, it has only $1.1$M parameters. For the loss function, we used the smooth $L1$-loss between the ground truth vector and the predicted vector, so that the model can learn the success rate distribution as closely as possible. To avoid overfitting, we made SPANet lightweight and used batch norm and early stopping.

Based on our previous human user study, we hypothesized that environmental position of the food relative to surrounding items would be a key factor in choosing an action.
SPANet, therefore, takes as input the surrounding environment type as one hot encoding of whether the food item is isolated, near a wall, or is stacked. It concatenates the encoding with the image feature vector from the base network.

\subsection{Environment Classifier Pipeline}
\begin{figure*}[t!]
  \includegraphics[width=\linewidth]{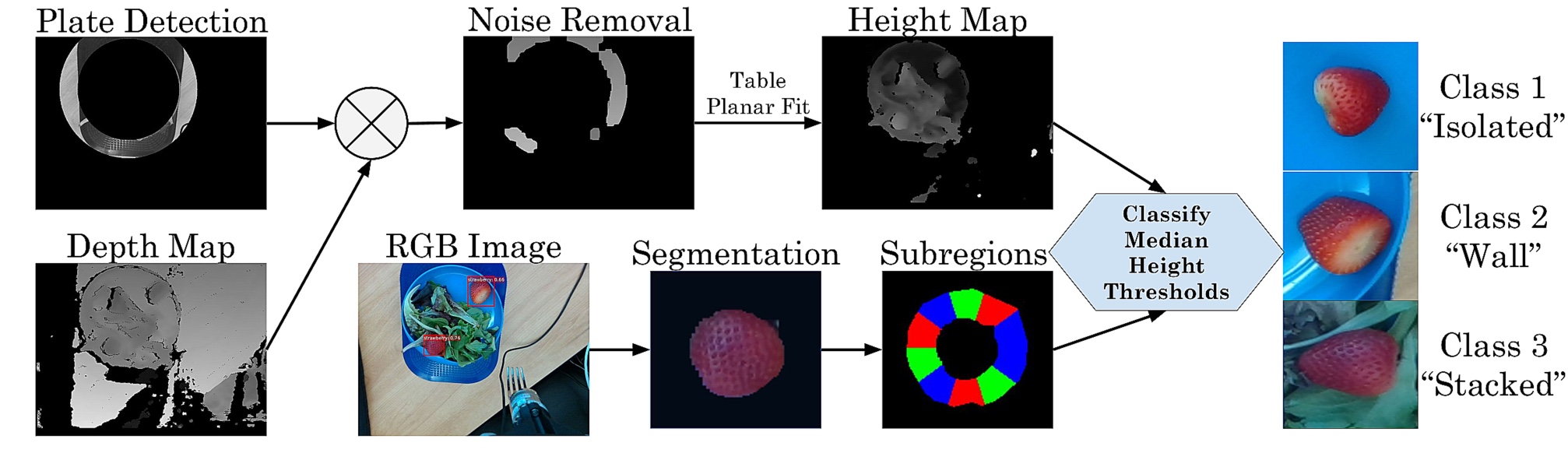}
 \caption{Environment classifier pipeline. A food item is categorized by comparing the depth of its surrounding environment with that of the table surface depth.}
 \label{fig:env_class}
\end{figure*}

Classifying the surrounding environment of a food item is not straightforward using SPANet itself because it requires looking at the cropped surroundings.
Therefore, we use a series of classical computer vision methods to extract this feature (see Fig. \ref{fig:env_class}). 

We first use a Hough circle transform to detect the plate and fit a planar table from the depth of the region immediately surrounding it. Subtracting the original depth map yields a height map relative to the table. After color-segmenting each food item from the RGB image, we divide the surrounding region of interest into sub-regions. If a sub-region intersects the plate boundary or if its median height exceeds some threshold, it is considered to be ``occupied.'' Otherwise, it is ``unoccupied.'' A food item is classified as ``isolated'' if a super-majority of the sub-regions are unoccupied, ``stacked'' if a super-majority of the sub-regions are occupied, and ``near-wall'' otherwise.

\subsection{Bite Acquisition Framework}
We developed a robotic feeding system that uses our framework to acquire food items from a plate by applying the actions defined in \secref{ssec:actions}. Note that our definition of \textit{action} specifies everything a motion planner needs to know in order to generate a trajectory, i.e., the approach angle and target pose. Given the success rates of each action on all visible items on the plate, the motion planner chooses a (target item, action) with the highest success rate and generates a trajectory. The robot executes it with a closed-loop, force-torque control until a certain force threshold is met or the trajectory has ended.

\section{Experiments}

\subsection{Experimental Setup}
Our setup consists of a 6 DoF JACO robotic arm~\cite{jaco}. The arm has 2 fingers that grab an instrumented fork (forque, see \figref{fig:schematic}) using a custom-built, 3D-printed fork holder. The system uses visual and haptic modalities to perform the feeding task. For haptic input, we instrumented the forque with a 6-axis ATI Nano25 Force-Torque sensor~\cite{ftsensor}. We use haptic sensing to control the end effector forces during skewering. For visual input, we mounted a custom built wireless perception unit on the robot's wrist; the unit includes the Intel RealSense D415 RGBD camera and the Intel Joule 570x for wireless transmission. Food is placed on an assistive bowl mounted on an anti-slip mat commonly found in assisted living facilities.

We experimented with 16 food items: apple, banana, bell pepper, broccoli, cantaloupe, carrot, cauliflower, celery, cherry tomato, grape, honeydew, kale, kiwi, lettuce, spinach and strawberry. 

\subsection{Data Collection}
For each food item, the robotic arm performed each of the six actions (see Fig. \ref{fig:schematic} right) under three different positional scenarios (see Fig. \ref{fig:schematic2}). For rotationally symmetrical items, such as kiwis, bananas, and leaves, the robot performed the trials with one fork roll. For each configuration (action, item, scenario), we collected 10 trials per item and marked success and failure rates.
Note that vision was not part of this data collection: we manually placed the food items directly under the fork such that when the fork hits the food, it would hit the center of the food item.
This manual placement was to ensure that the collected trials reflect how good an action is, assuming that the a perception module can correctly locate the center of the food and derive the fork to the desired location, as verified previously~\cite{gallenberger2019transfer}.
We defined \textit{failure} as at least 2 of 4 tines touching the food item but either the fork failing to skewer the item or the item falling off in less than 5 seconds.
If less than 2 fork tines touch a food item, we discard that trial because it indicates that the food item was misplaced.
We defined \textit{success} as the forque skewering the item and the item staying on the fork after being lifted up for 5 seconds. 

For each new trial, we changed the food item since every item has a slightly different size and shape even within the same class, and it is important to capture this variation.
During each skewering trial, we recorded RGBD images of the whole plate before and after acquisition, forces and torques and all joint positions of the arm during the trial, and whether the acquisition attempt was a success or a failure. 
Our entire dataset is available in the Harvard Dataverse~\cite{DVN/8A1XO3_2019}. 

\section{Dataset Analysis}
We analyzed the collected data to verify whether environmental features or the choice of actions with respect to the fork pitch and roll affect bite acquisition success rates. Effect of these variations would validate the need for the environment classifier as well as the choice of our action space.

We validated that the six actions and the three environment scenarios indeed resulted in different success rates. To test statistical significance, we performed a series of Fisher Exact tests for homogeneity as opposed to the more common t-test, u-test, or chi-squared test since our dataset was sparse. Our p-value threshold was 0.05 and, with Bonferroni correction, the corrected p-value threshold was 0.0024.

\begin{figure}[!t]
    \centering
    \begin{subfigure}[b]{0.32\columnwidth}
    \includegraphics[width=\linewidth]{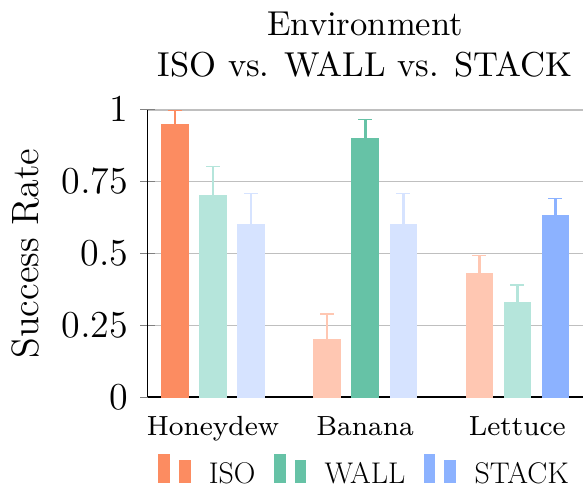}
    \label{fig:position}
    \end{subfigure}
    \begin{subfigure}[b]{0.32\columnwidth}
    \includegraphics[width=\linewidth]{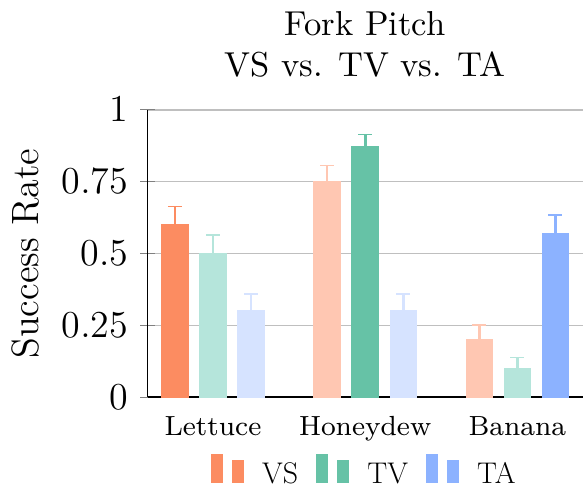}
    \label{fig:tilted-angled}
    \end{subfigure}
    \begin{subfigure}[b]{0.32\columnwidth}
    \includegraphics[width=\linewidth]{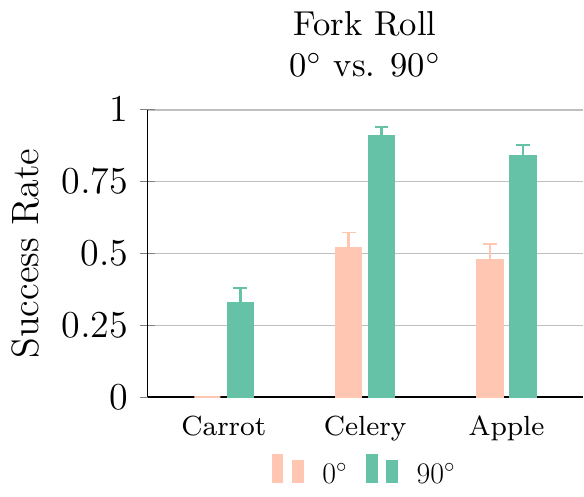}
    \label{fig:zero_ninety}
    \end{subfigure}
    \vspace{-0.5cm}
    \caption{\emph{Left: Surrounding environment affects the bite acquisition}: Even the same action on the same item has different success rates depending on how the item is placed with respect to other items or the plate. Vertical skewer on honeydew is the most successful when it is isolated. Tilted angled skewer on banana is a lot more successful when it is near wall. Any skewering action on lettuce is more successful when it is stacked. \emph{Center: Fork pitch affects bite acquisition}: Actions with different fork pitches perform differently for different food items. \emph{Right: Fork roll affects bite acquisition}: For long and slender items, aligning the fork tines perpendicular to the item ($90^\circ$) is more likely to succeed than horizontally.}
    \label{fig:environment-matters}
\end{figure}

\subsection{Surrounding Environment Affects Bite Acquisition}
We tested our three environment categories over all actions and food items. We found that the \emph{stacked} category played a significant role in bite acquisition, with a p-value of $0.0005$, compared to the wall or isolated categories. For these latter features, our current experiments did not find a statistically significant difference in success rates. Investigating further, we found that for a subset of food items -- viz., kiwi, banana, apple, bell pepper, broccoli, cantaloupe, cauliflower, and strawberry -- the \emph{wall} and \emph{stacked} environments are significantly better than the \emph{isolated} environment for the \emph{tilted-angled}~(TA) strategy, with a p-value of $0.0475$ and $0.0037$, respectively. This group was investigated because we empirically observed that these items needed a TA skewer to prevent their sliding off the fork tines, and that the TA skewer worked best near a wall or on top of items. However, note that the p-values exceed our corrected threshold.

\begin{figure}[!t]
    \centering
    \includegraphics[width=0.34\linewidth]{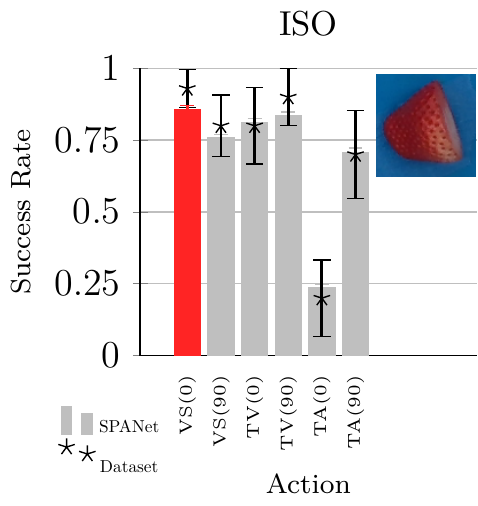}
    \includegraphics[width=0.31\linewidth]{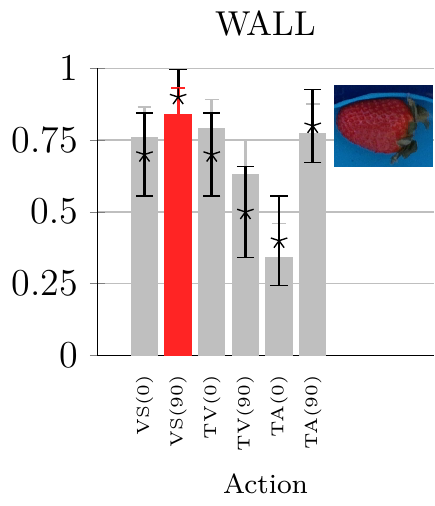}
    \includegraphics[width=0.31\linewidth]{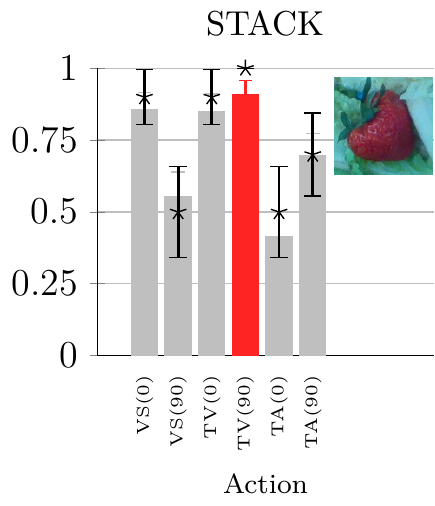}
    \vspace{-0.3cm}
    \caption{SPANet's action success rate predictions compared with the empirical success rates(*) for strawberry. Given an environment feature and a cropped image, SPANet successfully predicts the action distributions. More importantly, it correctly predicts the best actions (red).}
    \label{fig:action_dist}
    \vspace{-0.5cm}
\end{figure}

\figref{fig:environment-matters} left shows cases where the three environment scenarios result in different acquisition success rates for the same food item and action.
 \figref{fig:action_dist} further show that depending on the surrounding environment, a food item may have different success rate distributions and thus an intelligent bite acquisition system should decide its bite acquisition strategy not only based on a food item but also on its surrounding environment.

\subsection{Fork Pitch Affects Bite Acquisition}
We tested our three macro actions (which differ in their fork pitch angles) over all environment scenarios, food items, and fork roll angles. We found that the \emph{tilted-angled} (TA) action had a statistically different result, with a p-value of $0.0005$, from \emph{vertical-skewer} (VS) and \emph{tilted-vertical} (TV) actions. This result was echoed (albeit less significantly or not corrected) for a specific subset of food items in the stacked environment: kale, spinach, strawberry, kiwi, and honeydew (p-values of $0.0234$ and $0.0255$, respectively); for these items, we found that TV and VS outperformed TA. \figref{fig:environment-matters} center shows that different fork pitch actions work better for different food items.

\subsection{Fork Roll Affects Bite Acquisition}
We tested whether skewering a food perpendicular to the skewering axis results in a different success rate from skewering horizontally.
We did not find a strong significance, but since the test indicated a difference (p-value of $0.0030$) at our original p-value threshold, we hypothesize that with more data we could find a relationship. Investigating further, we found that for large long items with flat surfaces -- e.g., carrot, celery, and apple -- skewering at $90^\circ$ was better than at $0^\circ$ for all macro actions and environments. This implies that distinguishing these two fork rolls may be necessary for long items with flat surfaces. We compared carrot, apple, and celery and again found a p-value of $0.0044$, which is below our original threshold. This difference is echoed in \figref{fig:environment-matters} right. 

\begin{figure*}[t!]
\centering
\begin{subfigure}[b]{0.32\columnwidth}
   \includegraphics[width=\linewidth]{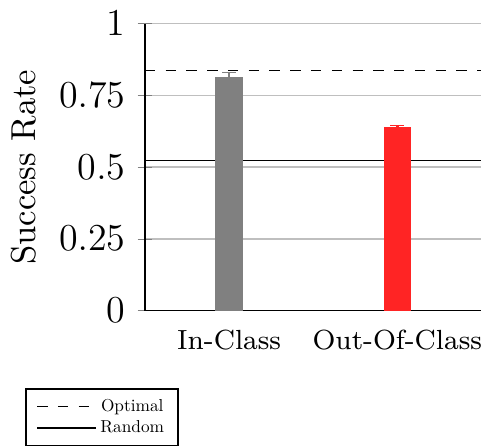}
   \vspace{-0.05cm}
 \end{subfigure}
 \begin{subfigure}[b]{0.67\columnwidth}
   \includegraphics[width=\linewidth]{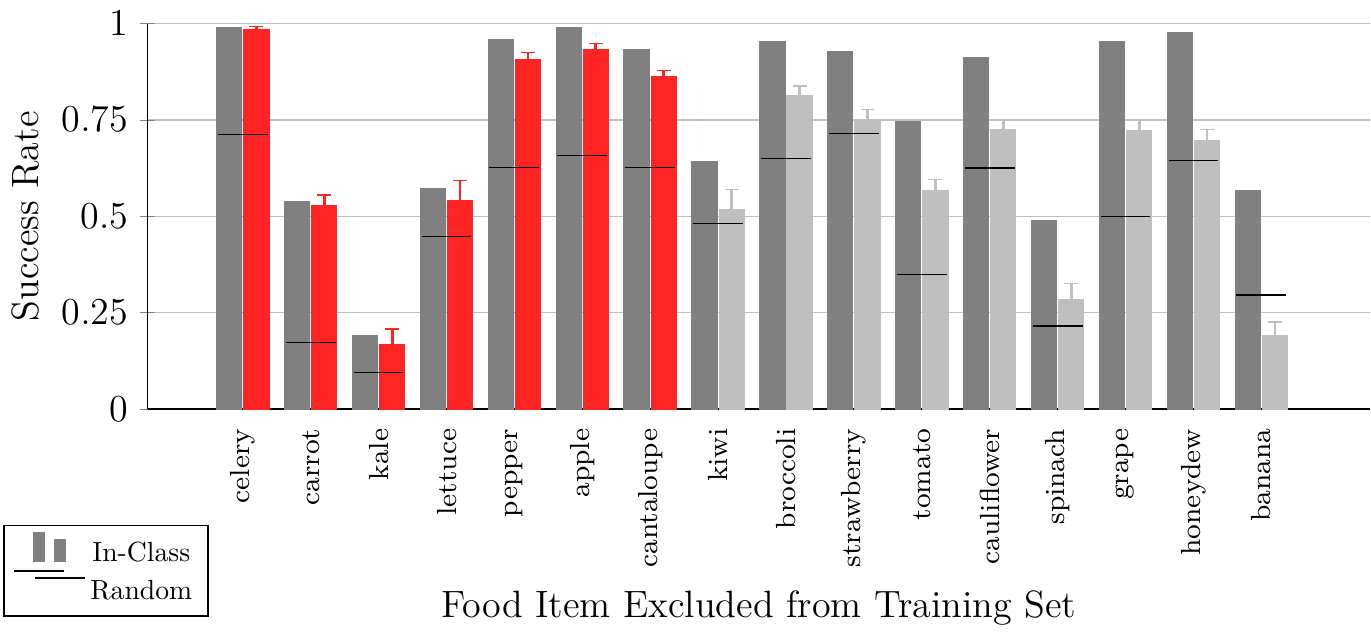}
 \end{subfigure}

 \caption{SPANet's generalizability to unseen food items (Out-Of-Class) compared with known food items (In-Class). Each bar represents the expected success rate of SPANet's proposal action, i.e., how well the best action proposed by SPANet would have performed according to the ground-truth data. \emph{Left:} Overall comparison. The success rate of SPANet's proposal action drops by 18\% for unseen food items.
 \emph{Right:} Per class comparison. For food classes from celery to cantaloupe, SPANet's OOC predictions stay very close to IC predictions. 
}
 \label{fig:results}
  \vspace{-0.3cm}
\end{figure*}

\section{Experimental Results}

In this section we discuss the training results of SPANet and full integration tests on the real robot system.
As an input to SPANet, we used images tightly cropped around each food item. Multiple images taken per trial from different angles to augment this dataset, leading to a total of approximately $10,000$ images.

\subsection{In-Class and Generalization Results}

\figref{fig:results} show SPANet's ability to generalize to food items that it has not seen during training (Out-Of-Class). SPANet generalizes well for 7 out of 16 items (the first 7 items in \figref{fig:results} right). Even when the gap is bigger (from kiwi to the right), SPANet's proposal action performs significantly better than a random action except for banana and kiwi, which have very different action distributions than the rest of the dataset.

Although the expected success rates even for the In-Class items are not 100\% for SPANet, it does not imply its inability to propose the right action. Even the ground-truth best action does not achieve 100\% success for most items. SPANet's proposal action has only 2.6\% lower success rate than the ground-truth best action across all food items when proposing for In-Class items, and it's 20\% lower for Out-Of-Class items. 

\begin{figure*}[t!]
\centering
 \begin{subfigure}[b]{0.4\columnwidth}
   \includegraphics[width=\linewidth]{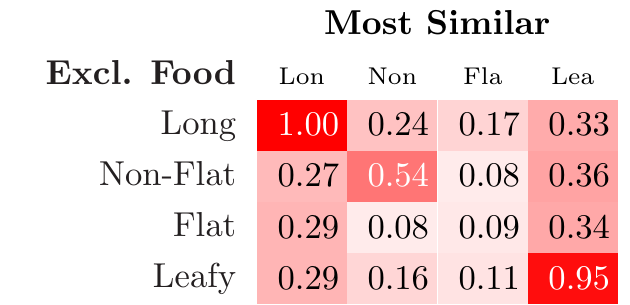}
   \vspace{0.65cm}
 \end{subfigure}
 ~
  \begin{subfigure}[b]{0.5\columnwidth}
   \includegraphics[width=0.48\linewidth]{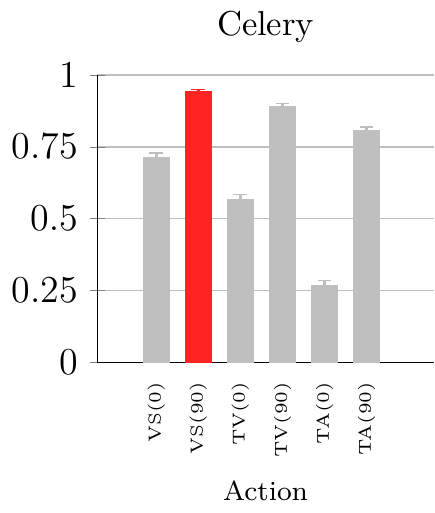}
   \hfill
   \includegraphics[width=0.48\linewidth]{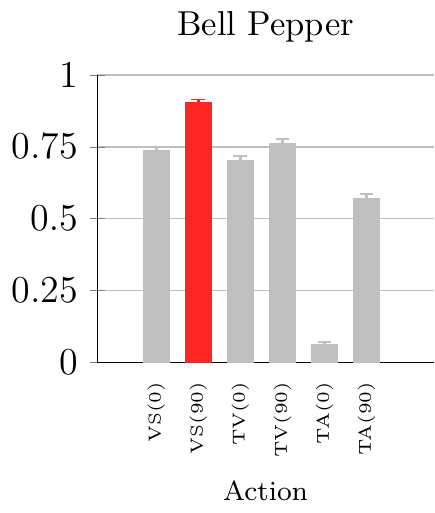}
 \end{subfigure}
 
 \caption{\emph{Left}: Normalized \emph{L2}-distance among SPANet's outputs on four categories of food items. SPANet makes similar predictions within ``long'' and ``leafy'' items \emph{Right}: Indeed, visually similar items such as celery and bell pepper have similar action success rate distributions, and their best actions are the same.}
 \label{fig:results_cluster}

\end{figure*}

To better understand how SPANet generalizes to unseen items, we compared the success rate predictions generated for Out-Of-Class~(OOC) items and In-Class~(IC) items. Figure \ref{fig:results_cluster} left shows the results clustered into four classes for easy visualization. We grouped food items into four categories based on visual properties, ``long", ``non-flat", ``flat", ``leafy". ``Leafy" included lettuce, spinach, kale, cauliflower, and broccoli. ``Long" included celery, carrots, apples, and bell pepper. ``Flat"  included cantaloupe and honeydew. ``Non-flat" included items with non-flat surfaces, such as strawberry, cherry tomato, and grape.  The results suggest that, given an unseen leafy item, SPANet's output is similar to other leafy items it has seen in the training set. The same can be said for long and non-flat items. We constructed this matrix by taking the expected success rate vector for each OOC food item and finding its nearest neighbor (under $L2$-distance) among the ground-truth of all food items. A small distance indicates that SPANet predicts the OOC item's action distribution to be similar to the nearest IC item. Indeed, some items have very similar success rate distributions, as in \figref{fig:results_cluster} right, a small $L2$ distance would be ideal. To mitigate a bias towards food items with success rate vectors near the mean of the whole dataset, we computed the softmax of the output vector before comparison. When grouping these results to generate \figref{fig:results_cluster} left, we took the maximum value within each group and normalized across all groups.

\subsection{Full System Integration Results}
To test the viability of SPANet on our bite acquisition framework, we integrated SPANet with our robotic system. We performed experiments with a representative set of foods and environments and measured the acquisition success rate. First, we tested carrot under all three environment scenarios to represent SPANet's ability to pick the correct angle for long and slender items. Second, we tested strawberry in the wall and stacked environment scenarios as a non-flat item where the best action depends on the environment. We collected 10 trials for each. To deal with positional accuracy issues with the fork being dislodged or bent during physical interaction with the environment, we used visual template matching to detect the fork-tip and adjust its expected pose to reduce image projection error. Anomalies external to the SPANet (see \secref{sec:discussion}), such as planning or wall detection/RetinaNet anomalies, were not counted towards the 10 trials.

\begin{figure*}[t!]
 \centering
  \includegraphics[width=0.24\linewidth]{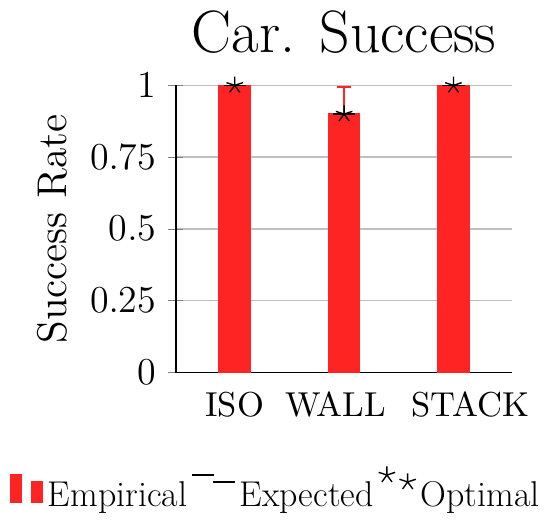}
  \includegraphics[width=0.24\linewidth]{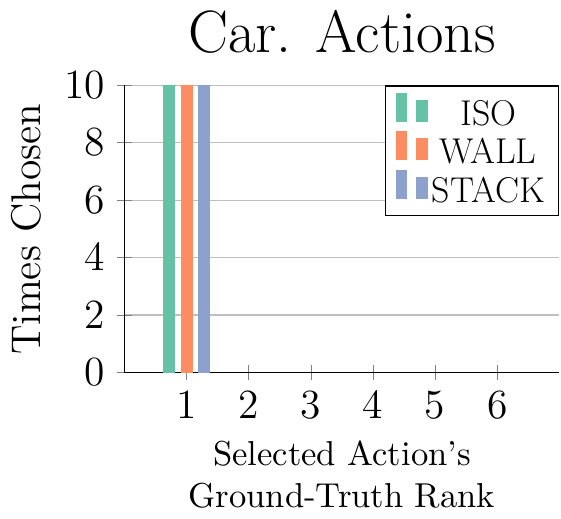}
  \hfill
  \includegraphics[width=0.24\linewidth]{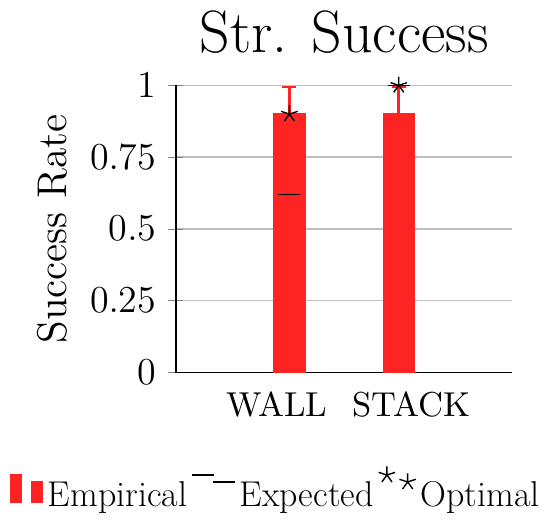}
  \includegraphics[width=0.24\linewidth]{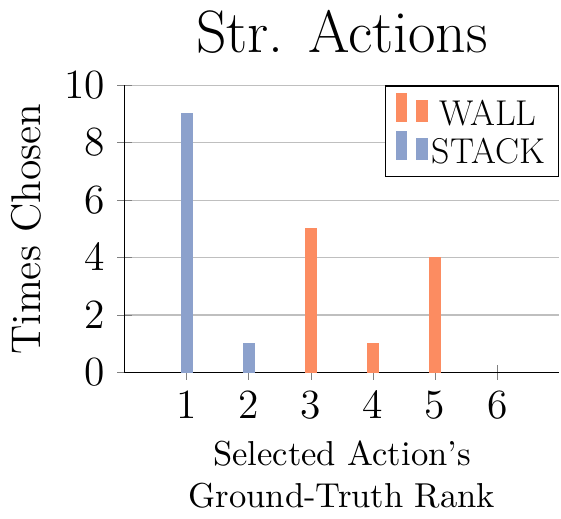}
 \caption{Full system integration acquisition success rates and selected actions for carrot \emph{(Left)} and strawberry \emph{(Right)}.
 }\label{fig:full_sys_success}
\end{figure*}

We found that SPANet tended to pick the most successful strategies to try to skewer carrots and strawberries. In each food-scenario pair except for strawberry-wall, SPANet always picked either the best or second best options. As shown in \figref{fig:full_sys_success}, the carrot tests perfectly matched their expected success rate and best action success rate. Strawberry-stacked experienced marginally less success than expected, having just 1 failure out of 10 where the fork went in and out without acquiring the strawberry. Interestingly, for strawberry-wall, the tests matched the best action success rate despite SPANet not picking the best actions in this case. This could perhaps be explained by slight variations in strawberry shapes and positions.

\subsection{Bite Acquisition from a Plate}

\begin{figure*}[t!]

\centering
 \begin{subfigure}[b]{0.34\linewidth}
  \includegraphics[width=\linewidth]{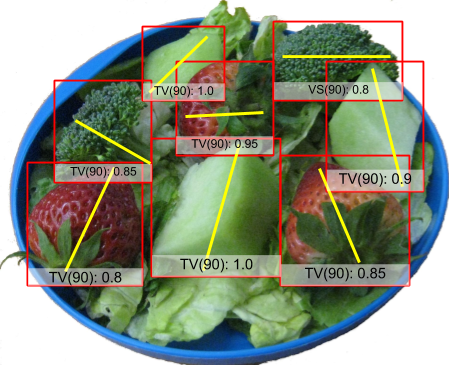}
 \caption{IC: 83\% Success}
 \end{subfigure}
 \begin{subfigure}[b]{0.31\linewidth}
  \includegraphics[width=\linewidth]{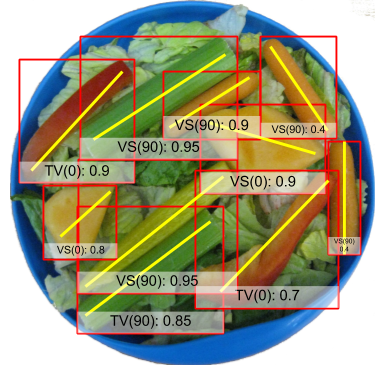}
 \caption{OOC1: 91\% Success}
 \end{subfigure}
 \begin{subfigure}[b]{0.31\linewidth}
  \includegraphics[width=\linewidth]{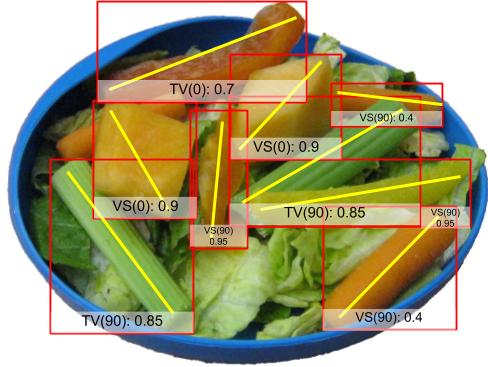}
 \caption{OOC2: 91\% Success}
 \end{subfigure}
   \caption{Food plates used for full integration tests, marked with best actions and their corresponding success rates predicted by SPANet. (a) Food plate with In-Class items resulted in 83\% success rate. (b) (c) Plates with Out-Of-Class items resulted in 91\% success rates. 
  The lower success rate of the IC plate is likely because some of the items are more challenging to acquire.}
  \label{fig:realistic_plate_results}
\end{figure*}

To demonstrate the viability of our bite acquisition system on a cluttered plate, we tested SPANet's ability to acquire a variety of foods from a cluttered full plate, as shown in \figref{fig:realistic_plate_results}. For these experiments, we trained a version of SPANet without a set of 3 food items (carrot, celery, and bell pepper). We tested two Out-Of-Class plates that contained all three excluded items plus cantaloupe in different configurations, as well as an In-Class plate containing honeydew, strawberry, and broccoli. We placed two to three pieces of each item onto a lettuce base, filling the plate so that items were often close to each other but not stacked on the lettuce, not themselves. Food item positions were chosen arbitrarily. We then attempted to acquire all food items on the plate except the lettuce.

As shown in \figref{fig:realistic_plate_results}, both Out-Of-Class plates had high success rates, acquiring all 10 items in only 11 attempts (excluding external anomalies, see \secref{sec:discussion}). Each attempt picked one of the best two actions. These results show that our system generalizes to unseen food items by picking reasonable actions and successfully acquiring items with few failures. The In-Class plate had lower performance. The first five items were picked up with only a single failure. The first honeydew attempt picked the best action. While the action choices for broccoli and strawberry were less consistent, they were acquired with only a single strawberry failure where the item slid off of the fork. We had a total of five planning failures where the system could not find a feasible motion plan to the commanded action. There were also perception anomalies with RetinaNet bounding box detection whereby two honeydew pieces and one broccoli piece could not be detected correctly due to the matching green background of lettuce.

\section{Discussion}\label{sec:discussion}

Our SPANet successfully generalized actions to unseen food items with similar action distributions as known ones. For soft and slippery items such as bananas or items with softer outside but harder core such as kiwis, the action distributions for successful bite acquisition significantly differed from the rest of the dataset, and hence our algorithm did not generalize well. This trait of slipperiness is difficult to capture by visual modality alone and is a topic of future work.

Compared to our previous work~\cite{gallenberger2019transfer} with a much larger action space, the reduced action space enables generalizability, but it remains to be seen if these $6$ actions would suffice for a wider range of food items. Additional actions could be added to our action space, including scooping or twirling. 

Our system's robustness could be improved to handle planning and perception anomalies. The system occasionally struggled to handle partially observable objects, leading to inaccurate box or axis detection. In terms of motion planning, some of the actions were more difficult to plan for when the item in clutter.This suggests a need for pre-skewer actions which maneuver food items to places where they can be skewered. Finally, as the fork holder is not rigidly attached to the robot's gripper, the fork tine's shape and position varied over time and led to positional variance at the end of the fork. While the fork-tip template's matching add-on helped, it would be beneficial to stabilize the fork and prevent it from bending out of shape due to repeated physical interactions with the environment.

\section*{ACKNOWLEDGMENT}
This work was funded by the National Institute of Health R01 (\#R01EB019335), National Science Foundation CPS (\#1544797), National Science Foundation NRI (\#1637748), the Office of Naval Research, the RCTA, Amazon, and Honda.


\bibliographystyle{IEEEtran}
\bibliography{skewering_generalization}

\end{document}